\documentclass[runningheads]{llncs}
\usepackage{graphicx}
\usepackage{amsfonts}
\usepackage{amsmath}
\usepackage[colorlinks=true]{hyperref}
\usepackage{cite}
\usepackage{multirow}
\usepackage{booktabs}
\usepackage{pifont}
\usepackage{makecell}
\usepackage{xcolor}
\usepackage[super]{nth}
\usepackage[misc,geometry]{ifsym}
\newcommand{\cmark}{\text{\ding{51}}}
\newcommand{\xmark}{\text{\ding{55}}}
\begin{document}
\title{Handwritten Mathematical Expression Recognition with Bidirectionally Trained Transformer}
\titlerunning{HMER with Bidirectionally Trained Transformer}
%
\author{Wenqi Zhao\inst{1}\orcidID{0000-0002-2952-0531} \and
Liangcai Gao\inst{1}(\Letter) \and
Zuoyu Yan\inst{1} \and
Shuai Peng\inst{1} \and
Lin Du\inst{2} \and
Ziyin Zhang\inst{2}}
\authorrunning{W. Zhao et al.}
%

\institute{Wangxuan Institute of Computer Technology, Peking University, Beijing, China \\
\email{1027572886a@gmail.com}\\
\email{\{gaoliangcai,yanzuoyu,pengshuaipku\}@pku.edu.cn} \and
Huawei AI Application Research Center\\
\email{\{dulin09,zhangziyin1\}@huawei.com}}

%
\maketitle              
\begin{abstract}
Encoder-decoder models have made great progress on handwritten mathematical expression recognition recently. However, it is still a challenge for existing methods to assign attention to image features accurately. Moreover, those encoder-decoder models usually adopt RNN-based models in their decoder part, which makes them inefficient in processing long \LaTeX{} sequences. In this paper, a transformer-based decoder is employed to replace RNN-based ones, which makes the whole model architecture very concise. Furthermore, a novel training strategy is introduced to fully exploit the potential of the transformer in bidirectional language modeling. Compared to several methods that do not use data augmentation, experiments demonstrate that our model improves the ExpRate of current state-of-the-art methods on CROHME 2014 by 2.23\%. Similarly, on CROHME 2016 and CROHME 2019, we improve the ExpRate by 1.92\% and 2.28\% respectively.

\keywords{handwritten mathematical expression recognition \and transformer \and bidirection \and encoder-decoder model}
\end{abstract}
\section{Introduction}
The encoder-decoder models have shown quite effective performance on various tasks such as scene text recognition~\cite{cheng2017focusing} and image captioning~\cite{xu2015show}. Handwritten Mathematical Expression Recognition (HMER) aims to generate the math expression \LaTeX{} sequence according to the handwritten math expression image. Since HMER is also an image to text modeling task, many encoder-decoder models~\cite{zhang2017watch, zhang2018multi} have been proposed for it in recent years.

However, existing methods suffer from the lack of coverage problem~\cite{zhang2017watch} to varying degrees. This problem refers to two possible manifestations: over-parsing and under-parsing. Over-parsing means that some regions of HME image are redundantly translated multiple times, while under-parsing denotes that some regions remain untranslated.

Most encoder-decoder models are RNN-based models, which have difficulty modeling the relationship between two symbols that are far apart. Previous study~\cite{bengio1993problem} had noted this long-term dependency problem caused by gradient vanishing. This problem is exposed more obviously in HMER task. Compared to traditional natural language processing, \LaTeX{} is a markup language designed by human, and thus has a clearer and more distinct syntactic structure, e.g., ``\{" and ``\}" are bound to appear in pairs. When dealing with long \LaTeX{} sequences, it is difficult for RNN-based models to capture the relationship between two distant ``\{" and ``\}" symbol, resulting in lack of awareness of the \LaTeX{} syntax specification.

Traditional autoregressive models~\cite{zhang2017watch, zhang2018multi} use left-to-right (L2R) direction to predict symbols one by one in the inference phase. Such approaches may generate unbalanced outputs~\cite{liu2016agreement}, which prefixes are usually more accurate than suffixes. To overcome this problem, existing study~\cite{liu2016agreement} employ two independent decoders, trained for left-to-right and right-to-left directions, respectively. This usually leads to more parameters and longer training time. Therefore, an intuitive attempt is to adapt a single decoder for bi-directional language modeling.

In this paper, we employ the transformer~\cite{vaswani2017attention} decoder into HMER task, alleviating the lack of coverage problem~\cite{zhang2017watch} by using positional encodings. Besides, a novel bidirectional training strategy is proposed to obtain a \textbf{B}idirectionally \textbf{T}rained \textbf{TR}ansformer (BTTR) model. The strategy enables single transformer decoder to perform both L2R and R2L decoding. We further show that our BTTR model outperforms RNN-based ones in terms of both training parallelization and inferencing accuracy. The main contributions of our work are summarized as follows:
\let\labelitemi\labelitemii
\begin{itemize}
  \item To the best of our knowledge, it is the first attempt to use end-to-end trained transformer decoder for solving HMER task.
  \item The combination of image and word positional encodings enable each time step to accurately assign attention to different regions of input image, alleviating the lack of coverage problem.
  \item A novel bidirectional training strategy is proposed to perform bidirectional language modeling in a single transformer decoder.
  \item Compared to several methods that do not use data augmentation,  experiments demonstrate that our method obtains new SOTA performance on various dataset, including an ExpRate of 57.91\%,  54.49\%, and 56.88\% on the CROHME 2014~\cite{mouchere2014icfhr}, CROHME 2016~\cite{mouchere2016icfhr2016}, and CROHME 2019~\cite{mahdavi2019icdar} test sets, respectively.
  \item We make our code available on the GitHub.\footnote{\url{https://github.com/Green-Wood/BTTR}}
\end{itemize}

\section{Related Work}

\subsection{HMER Methods}
In the last decades, many approaches~\cite{chan2000mathematical, zanibbi2012recognition, zhang2017symbol, yan2021convmath, jiang2018mathematics, yuan2016mathematical, yuan2018formula} related to HMER have been proposed. These approaches can be divided into two categories: grammar-based and encoder-decoder based. In this section, we will briefly review the related work in both categories.

\subsubsection{Grammar Based}
These methods usually consist of three parts: symbol segmentation, symbol recognition, and structural analysis. Researchers have proposed a variety of predefined grammars to solve HMER task, such as stochastic context-free grammars~\cite{alvaro2014recognition}, relational grammars~\cite{maclean2013new}, and definite clause grammars~\cite{chan2001error, chan2000efficient}. None of these grammar rules are data-driven, but are hand-designed, which could not benefit from large dataset.

\subsubsection{Encoder-Decoder Based}
In recent years, a series of encoder-decoder models have been widely used in various tasks~\cite{yuan2020automatic,yuan2020follow, yan2021persistence}. In HMER tasks, Zhang et al.~\cite{zhang2017watch} observed the lack of coverage problem and proposed WAP model  to solve the HMER task. In the subsequent studies, DenseWAP~\cite{zhang2018multi} replaced VGG encoder in the WAP with DenseNet~\cite{huang2017densely} encoder, and improved the performance. Further, DenseWAP-TD~\cite{zhang2020treedecoder} enhanced the model's ability to handle complex formulas by substituting string decoder with a tree decoder. Wu et al.~\cite{wu2021graph-to-graph} used stroke information and formulated the HMER as a graph-to-graph(G2G) modeling task. Such encoder-decoder based models have achieved outstanding results in several CROHME competitions~\cite{mouchere2014icfhr, mouchere2016icfhr2016, mahdavi2019icdar}.

\subsection{Transformer}
Transformer~\cite{vaswani2017attention} is a neural network architecture based solely on attention mechanisms. Its internal self-attention mechanism makes transformer a breakthrough compared to RNN in two aspects. Firstly, transformer does not need to depend on the state of the previous step as RNN does. Well-designed parallelization allows transformer to save a lot of time in the training phase. Secondly, tokens in the same sequence establish direct one-to-one connections through the self-attention mechanism. Such a mechanism fundamentally solves the gradient vanishing problem of RNN~\cite{bengio1993problem}, making transformer more suitable than RNN on long sequences. In recent years, RNN is replaced by transformer in various tasks in computer vision and natural language processing~\cite{parmar2018image, devlin2018bert}.

Recently, transformer has been used in offline handwritten text recognition. Kang et al.~\cite{kang2020pay} first adopted transformer networks for the handwritten text recognition task and achieved state-of-the-art performance. For the task of mathematical expression, ``Univ. Linz" method in CROHME 2019~\cite{mahdavi2019icdar} used a Faster R-CNN detector and a transformer decoder. Two subsystems were trained separately.

\subsection{Right-to-Left Language Modeling}
To solve the problem that traditional autoregressive models can only perform left-to-right(L2R) language modeling, many studies have attempted right-to-left(R2L) language modeling. Liu et al.~\cite{liu2016agreement} trained a R2L model separately. During the inference phase, hypotheses from L2R and R2L models are re-ranked to produce the best candidate. Furthermore, Zhang et al.~\cite{zhang2019regularizing} used R2L model to regularize L2R model in the training phase to obtain a better L2R model. Zhou et al.~\cite{zhou2019synchronous} proposed SB-NMT that utilized a single decoder to generate sequences bidirectionally. However, all of these methods increase model complexity. On the other hand, our approach achieves bidirectional language modeling on a single decoder while keeping the model concise.

\section{Methodology}
\begin{figure}
	\centering
	\includegraphics[scale=0.32]{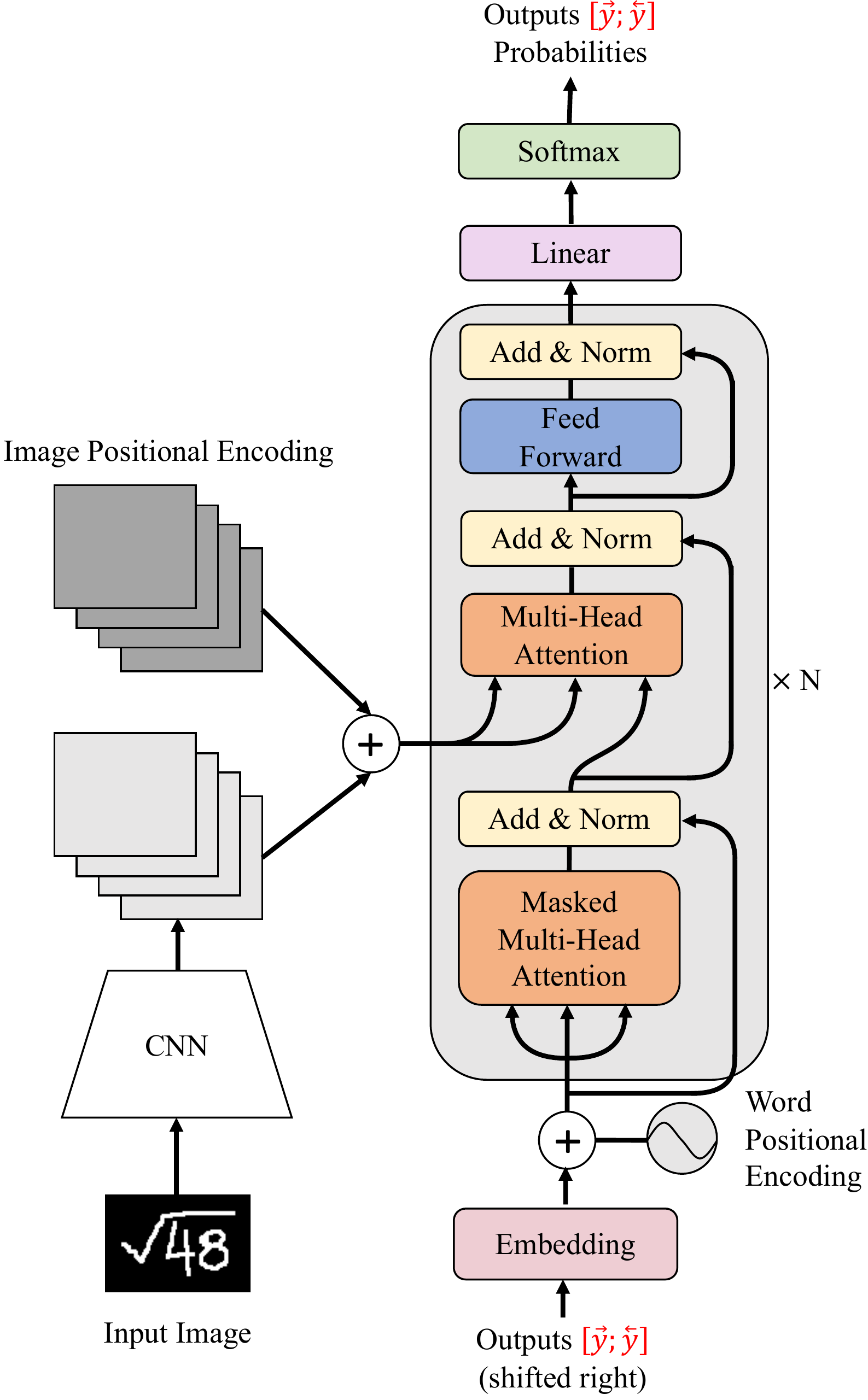}
	\caption{The architecture of BTTR model. L2R and R2L sequences $\color{red} [\protect\overrightarrow{y};\protect\overleftarrow{y}]$ are concatenated through the batch dimension as the input to decoder part.}
	\label{fig:model}
\end{figure}

In this section, we will detailed introduce the proposed BTTR model architecture, as illustrated in Fig.~\ref{fig:model}. In section~\ref{sec:cnn}, we will briefly describe the DenseNet~\cite{huang2017densely} model used in the encoder part. In section~\ref{sec:pos} and~\ref{sec:decoder}, the positional encodings and transformer model used in the encoder and decoder part will be described in detail. Finally in section~\ref{sec:bi}, we will introduce the proposed novel bidirectional training strategy, which allows to perform bidirectional language modeling in a single transformer decoder.

\subsection{CNN Encoder} \label{sec:cnn}
In the encoder part, DenseNet is used as the feature extractor for HME images. The main idea of DenseNet is to increase information flow between layers by introducing direct connections between each layer and all its subsequent layers. In this way, given output features $\mathbf{x}_{0}, \mathbf{x}_{1}, \ldots, \mathbf{x}_{l-1}$ from $0^{th}$ to $(l-1)^{th}$ layer, the output feature of $l^{th}$ layer can be computed by:
\begin{equation}
	\mathbf{x}_{\ell}=H_{\ell}\left(\left[\mathbf{x}_{0}; \mathbf{x}_{1}; \ldots; \mathbf{x}_{\ell-1}\right]\right)
\end{equation}
where $\left[\mathbf{x}_{0}; \mathbf{x}_{1}; \ldots; \mathbf{x}_{\ell-1}\right]$ denotes the concatenation operation of all the output features, and $H_{\ell}(\cdot)$ denotes a composite function of three consecutive layers: a batch normalization (BN)~\cite{ioffe2015batch} layer, followed by a ReLU~\cite{glorot2011deep} layer and a $3 \times 3$ convolution (Conv) layer. 

Through concatenation operation in the channel dimension, DenseNet enables better propagation of gradient. The paper~\cite{huang2017densely} states that by this dense connection, DenseNet can achieve better performance with fewer parameters compared to ResNet~\cite{he2016deep}.

In addition to DenseNet, we also add a $1 \times 1$ convolution layer in the encoder part to adjust the image feature dimension to the size of embedding dimension $d_{\text{model}}$ for subsequent processing.

\subsection{Positional Encoding} \label{sec:pos}
The positional information of image features and word vectors can effectively help the model to identify regions that need to attend. In the previous studies~\cite{zhang2017watch, zhang2018multi}, although the RNN-based model inherently takes the order of word vectors into account, it neglects the positional information of image features. 

In this paper, since the transformer model itself doesn't have any sense of position for each input vector, we use two types of positional encodings to address this information. In detail, we use image positional encodings and word positional encodings to represent the image feature position and word vector position, respectively. 

We refer to image features and word vector features as \textit{content-based}, and the two types of positional encodings as \textit{position-based}. As illustrated in Fig.~\ref{fig:model}, \textit{content-based} and \textit{position-based} features are summed up, as the input to transformer decoder.

\subsubsection{Word Positional Encoding}
Word positional encoding is basically the same as sinusoidal positional encoding proposed in the original transformer work~\cite{vaswani2017attention}. Given position $pos$ and dimension $d$ as input, the word positional encoding vector $\mathbf{p}^{W}_{pos, d}$ is defined as:
\begin{equation}
	\mathbf{p}^{W}_{p o s, d}[2 i]=\sin (p o s / 10000^{2 i / d})
\end{equation}
\begin{equation}
	\mathbf{p}^{W}_{p o s, d}[2 i + 1]=\cos (p o s / 10000^{2 i / d})
\end{equation}
where $i$ is the index in dimension.

\subsubsection{Image Positional Encoding}
A 2-D normalized positional encoding is used to represent the image position features. We first compute sinusoidal positional encoding $\mathbf{p}^{W}_{p o s, d/2}$ in each of the two dimensions and then concatenate them together. Given a 2-D position tuple $(x, y)$ and the same dimension $d$ as the word positional encoding, the image positional encoding vector $\mathbf{p}^{I}_{x, y, d}$ is represented as:
\begin{equation}
	\bar{x} = \frac{x}{H}, \quad \bar{y} = \frac{y}{W}
\end{equation}
\begin{equation}
	\mathbf{p}^{I}_{x,y,d} = [\mathbf{p}^{W}_{\bar{x}, d/2}; \mathbf{p}^{W}_{\bar{y}, d/2}]
\end{equation}
where $H$ and $W$ are height and width of input images.

\subsection{Transformer Decoder} \label{sec:decoder}
For the decoder part, we use the standard transformer model~\cite{vaswani2017attention}. Each basic transformer decoder layer module consists of four essential parts. In the following, we will describe the implementation details of these components.

\subsubsection{Scaled Dot-Product Attention}
This attention mechanism is essentially using the query to obtain the value from key-value pairs, based on the similarity between the query and key. The output matrix can be computed in parallel by the query $\mathbf{Q}$, key $\mathbf{K}$, value $\mathbf{V}$, and dimension $d_k$.
\begin{equation}
	\operatorname{Attention}(\mathbf Q, \mathbf K, \mathbf V)=\operatorname{softmax}(\frac{\mathbf Q \mathbf K^{T}}{\sqrt{d_{k}}}) \mathbf V
\end{equation}

\subsubsection{Multi-Head Attention}
With multi-head mechanism, the scaled dot-product attention module can attend to feature-map from multiple representation subspaces jointly. With projection parameter matrices 
$
\mathbf W_{i}^{Q} \in \mathbb{R}^{d_{\text {model }} \times d_{k}}, 
\mathbf W_{i}^{K} \in \mathbb{R}^{d_{\text {model }} \times d_{k}}, 
\mathbf W_{i}^{V} \in \mathbb{R}^{d_{\text {model }} \times d_{v}}
$
, we first project the query $\mathbf{Q}$, key $\mathbf{K}$, and value $\mathbf{V}$ into a subspace to compute the head $\mathbf{H}_i$.
\begin{equation}
\mathbf{H}_i=\operatorname{Attention}\left(\mathbf Q \mathbf W_{i}^{Q}, \mathbf K \mathbf W_{i}^{K},\mathbf V \mathbf W_{i}^{V}\right)
\end{equation}
Then all the heads are concatenated and projected with a parameter matrix
$
\mathbf W^{O} \in \mathbb{R}^{h d_{v} \times d_{\text {model }}}
$
and the number of heads $h$.
\begin{equation}
	\operatorname{MultiHead}(\mathbf Q, \mathbf K, \mathbf V)=\left[\mathbf H_1; \ldots; \mathbf H_{h}\right] \mathbf W^{O}
\end{equation}

\subsubsection{Masked Multi-Head Attention}
In the decoder part, due to the autoregressive property, the next symbol is predicted based on the input image and previously generated symbols. In the training phase, a lower triangle mask matrix is used to enable the self-attention module to restrict the attention region for each time step. Due to masked multi-head attention mechanism,  the whole training process requires only one forward computation.

\subsubsection{Position-wise Feed-Forward Network}
Feed-Forward Network(FNN) consists of three operations: a linear transformation, a ReLU activation function and another linear transformation. 
\begin{equation}
	\operatorname{FFN}(\mathbf x)=\max \left(0, \mathbf x \mathbf W_{1}+\mathbf b_{1}\right) \mathbf W_{2} + \mathbf b_{2}
\end{equation} 
After multi-head attention, the information between positions has been fully exchanged. FFN enables each position to integrate its own internal information separately.

\subsection{Bidirectional Training Strategy} \label{sec:bi}
First, two special symbols ``$\langle \text{SOS} \rangle$" and ``$\langle \text{EOS} \rangle$" are introduced in the dictionary to denote the start and end of the sequence. For the target \LaTeX{} sequence $y=\{ y_{1}, \ldots, y_{T}\}$, we denote the target sequence from left to right (L2R) as $\overrightarrow{y} =\{\langle \text{SOS} \rangle, y_{1}, \ldots, y_{T}, \langle \text{EOS} \rangle\}$, and the right-to-left (R2L) target sequence as $\overleftarrow{y} =\{\langle \text{EOS} \rangle, y_{T}, \ldots, y_{1}, \langle \text{SOS} \rangle\}$.

Conditioned on image $x$ and model parameter $\theta$, the traditional autoregressive model need to compute the probability distribution:
\begin{equation} \label{eq:l2robj}
	p\left(\overrightarrow{y}_{j} \mid \overrightarrow{y}_{<j}, x, \theta\right)
\end{equation}
where $j$ is the index in target sequence.

In this paper, since the transformer model itself does not actually care about the order of input symbols, we can use a single transformer decoder for bi-directional language modeling. Modeling both Eq.~\eqref{eq:l2robj} and Eq.~\eqref{eq:r2lobj} at the same time.
\begin{equation} \label{eq:r2lobj}
	p\left(\overleftarrow{y}_{j} \mid \overleftarrow{y}_{<j}, x, \theta\right)
\end{equation}
To achieve this goal, a simple yet effective bidirectional training strategy is proposed, in which for each training sample, we generate two target sequences, L2R and R2L, from the target \LaTeX{} sequence, and compute the training loss in the same batch (details in Section~\ref{sec:training}). Compared with unidirectional language modeling, our approach trains a model to perform bidirectional language modeling without sacrificing model conciseness. Experiment results in Section~\ref{sec:bt_eval} verify the effectiveness of our bidirectional training strategy.

\section{Implementation Details}
\subsection{Networks}
In the encoder part, to make a fair comparison with the previous state-of-the-art method, we use the same DenseNet feature extractor as the DenseWAP model~\cite{zhang2018multi}. Specifically, three bottleneck layers are used in the backbone network and transition layers are added in between to reduce the number of feature-maps. In each bottleneck layer, we set the growth rate to $k=24$, the depth of each block to $D=16$, and the compression hyperpatameter of the transition layer to $\theta=0.5$.

In the decoder part, we use the standard transformer model. We set the embedded dimension and model dimension to $d_{\text{model}}=256$, the number of heads in the multi-head attention module to $H=8$, the dimension of intermediate layers in the FFN to $d_{ff} = 1024$, and the number of transformer decoder layer to $N=3$. The 0.3 dropout rate is used to prevent overfitting.

\subsection{Training} \label{sec:training}
Our training objective is to maximize the predicted probability of the ground truth symbols in Eq.~\eqref{eq:l2robj} and Eq.~\eqref{eq:r2lobj}, so we use the standard cross-entropy loss function to calculate the loss between the predicted probabilities w.r.t. the ground truth at each decoding position. Given the training sample $\left\{x^{(z)}, y^{(z)}\right\}_{z=1}^{Z}$, the objective function for optimization is shown as follows:
\begin{equation}
	\overrightarrow{\mathcal{L}}_{j}^{(z)}(\theta) = -\log p(\overrightarrow{y}_{j}^{(z)} \mid \overrightarrow{y}_{<j}^{(z)}, x^{(z)},\theta) 
\end{equation}
\begin{equation}
	\overleftarrow{\mathcal{L}}_{j}^{(z)}(\theta) = -\log p(\overleftarrow{y}_{j}^{(z)} \mid \overleftarrow{y}_{<j}^{(z)}, x^{(z)},\theta) 
\end{equation}
\begin{equation}
	\mathcal{L}(\theta) = \frac{1}{2ZL} \sum_{z=1}^{Z} \sum_{j=1}^{L}\left(\overrightarrow{\mathcal{L}}_{j}^{(z)}(\theta) + \overleftarrow{\mathcal{L}}_{j}^{(z)}(\theta) \right)
\end{equation}
The model is trained from scratch using the Adadelta algorithm~\cite{zeiler2012adadelta} with a weight decay of $10^{-4}$ , $\rho=0.9$, and $\epsilon=10^{-6}$. PyTorch framework is used to implement our model. The model is trained on four NVIDIA 1080Ti GPUs with $11 \times 4$ GB memory.

\subsection{Inferencing}
In the inference phase, we aim to generate the most likely \LaTeX{} sequence conditioned on the input image. Which can be fomulated as follows:
\begin{equation}
	\hat{\mathbf{y}}=\underset{\mathbf{y}}{\operatorname{\mathbf{argmax}}} \ p\left(\mathbf{y} \mid \mathbf{x}, \theta\right)
\end{equation}
where $\mathbf{x}$ denotes the input image and $\theta$ denotes the model parameter.

Unlike the training phase where a lower triangular mask matrix is used to generate the prediction for all time steps simultaneously. Since we have no ground truth of the previously predicted symbol, we can only predict symbols one by one until the ``End" symbol appears or the predefined maximum length is reached.

Obviously, we cannot search for all possible sequences, thus a heuristic beam search is proposed to balance the computational cost with the quality of decoding. Further, taking advantage of the fact that our decoder is capable of bidirectional language modeling, approximate joint search~\cite{liu2016agreement} is used to improve the performance. The basic idea consists of three steps: (1) Firstly, a beam search is performed on L2R and R2L directions to obtain two k-best hypotheses.(2) Then, we reverse L2R hypotheses to R2L direction and R2L hypotheses to L2R direction and treat these hypotheses as ground truth to compute the loss values for each of them as in the training phase.(3) Finally, those loss values are added to their original hypothesis scores to obtain the final scores, which is then used to find the best candidate. In practice, we set beam size $k=10$, the maximum length to 200, and length penalty $\alpha=1.0$.

\section{Experiments}
\subsection{Datasets}
We use the Competition on Recognition of Online Handwritten Mathematical Expressions (CROHME) benchmark, which currently is the largest dataset for handwritten mathematical expression to validate the proposed model. We use the same training dataset but different test datasets. The training set contains 8836 handwritten mathematical expressions, while the test sets of CROHME 2014/2016/2019 contain 986/1147/1199 expressions respectively.

In the CROHME dataset, each handwritten math expression is saved in a InkML file, which contains handwritten stroke trajectory information and ground truth in both MathML and \LaTeX{} formats. We transform the handwritten stroke trajectory information in the InkML files to offline images in bitmap format for training and testing. With official evaluation tools provided by the CROHME 2019~\cite{mahdavi2019icdar} organizers and the ground truth in symLG format, we convert the predicted \LaTeX{} sequences to symLG format and evaluate the performance.

\subsection{Compare with state-of-the-art results}
The results of some models on the CROHME 2014/2016/2019 datasets are shown in Table~\ref{tb:sota}. To ensure the fairness of performance comparison, the methods we show all use only the officially provided 8836 training samples. Neither we nor the methods we compare use data augmentation.

We first provide results of three traditional handwritten mathematical formula recognition methods based on tree grammar in CROHME 2014 as the baseline, denoted as \uppercase\expandafter{\romannumeral1}, \uppercase\expandafter{\romannumeral6}, and \uppercase\expandafter{\romannumeral7}. For CROHME 2016, we provide the best performing ``TOKYO" method as the baseline, which only used official training samples. For CROHME 2019 official methods, we provide ``Univ. Linz"~\cite{mahdavi2019icdar} method as the baseline. For Image-to-\LaTeX{} methods, we use the previous state-of-the-art ``WYGIWYS"~\cite{deng2016you}, ``PAL-v2"~\cite{wu2020handwritten}, ``WAP"~\cite{zhang2017watch}, ``Weakly supervised WAP"(WS WAP)~\cite{truong2020improvement}, ``DenseWAP"~\cite{zhang2018multi} as well as the tree decoder-based ``DenseWAP-TD"~\cite{zhang2020treedecoder} method. The ``Ours-Uni" and ``Ours-Bi" methods denote the model trained using the vanilla transformer decoder and the model trained with the bidirectional training strategy on top of it.

In Table~\ref{tb:sota}, by comparing ``DenseWAP" with ``Ours-Uni", both are unidirectional string decoder based models, we can obtain 4.29\% average performance improvement in ExpRate by simply replacing the RNN-based decoder with the vanilla transformer decoder. 

Compared with other methods, our proposed BTTR model outperforms the previous state-of-the-art methods in nearly all metrics and is about 3.4\% ahead of the ``DenseWAP-TD" method in ExpRate, which explicitly encodes our prior knowledge about the \LaTeX{} grammar through a tree decoder.

\begin{table}[htbp]
\renewcommand\arraystretch{1.1}
\caption{Performance comparison of single models on the CROHME 2014/2016/2019 test sets(in \%), where ``ExpRate", ``$\leq \text{1 error}$" and ``$\leq \text{2 error}$" columns mean expression recognition rate when zero to two structural or symbol errors can be tolerated. ``StruRate" column means structure recognition rate.}
\begin{center}
\begin{tabular}{p{0.2\textwidth}<{\centering}ccp{0.15\textwidth}<{\centering}p{0.15\textwidth}<{\centering}p{0.15\textwidth}<{\centering}}
\toprule
Dataset & Model & ExpRate & $\leq \text{1 error}$ & $\leq \text{2 error}$ & StruRate\\
\bottomrule
\multirow{11}{*}{CROHME14}
& \uppercase\expandafter{\romannumeral1} & 37.22 & 44.22 & 47.26 & - \\
& \uppercase\expandafter{\romannumeral6} & 25.66 & 33.16 & 35.90 & - \\
& \uppercase\expandafter{\romannumeral7} & 26.06 & 33.87 & 38.54 & - \\
& WYGIWYS & 36.4 & - & - & - \\
& WAP & 40.4 & 56.1 & 59.9 & - \\
& DenseWAP & 43.0 & 57.8 & 61.9 & 63.2 \\
& PAL-v2 & 48.88 & 64.50  & 69.78  & - \\
& DenseWAP-TD & 49.1 & 64.2 & 67.8 & 68.6 \\
& WS WAP & 53.65 & - & - & - \\
\cline{2-6}
& Ours-Uni & 48.17 & 59.63 & 63.29 & 65.01 \\
& \textbf{Ours-Bi} & \textbf{53.96} & \textbf{66.02} & \textbf{70.28} & \textbf{71.40} \\
\bottomrule
\multirow{8}{*}{CROHME16}
& TOKYO & 43.94 & 50.91 & 53.70 & 61.6 \\
& WAP & 37.1 & - & - & - \\
& DenseWAP & 40.1 & 54.3 & 57.8 & 59.2 \\
& PAL-v2  & 49.61  & 64.08  & \textbf{70.27} & -  \\
& DenseWAP-TD & 48.5 & 62.3 & 65.3 & 65.9 \\
& WS WAP & 51.96 & \textbf{64.34} & 70.10 & - \\
\cline{2-6}
& Ours-Uni & 44.55 & 55.88 & 60.59 & 61.55 \\
& \textbf{Ours-Bi} & \textbf{52.31} & 63.90 & 68.61 & \textbf{69.40} \\
\bottomrule
\multirow{5}{*}{CROHME19}
& Univ. Linz & 41.49 & 54.13 & 58.88 & 60.02 \\
& DenseWAP & 41.7 & 55.5 & 59.3 & 60.7 \\
& DenseWAP-TD & 51.4 & \textbf{66.1} & 69.1 & 69.8 \\
\cline{2-6}
& Ours-Uni & 44.95 & 56.13 & 60.47 & 60.63 \\
& \textbf{Ours-Bi} & \textbf{52.96} & 65.97 & \textbf{69.14} & \textbf{70.06} \\
\bottomrule
\end{tabular}
\label{tb:sota}
\end{center}
\end{table}

\subsection{Ablation Study} \label{sec:bt_eval}
In Table~\ref{tb:ablation}, the first ``IPE" column denotes whether to use image positional encoding or not. Secondly, the ``Bi-Trained" column shows whether the bidirectional training strategy is used. On the ``AJS" column, $\cmark$ indicates the use of approximate joint search~\cite{liu2016agreement}, while $\xmark$ represents L2R search. The last ``Ensemble" column denotes whether the ensemble method is used.

First, we can see that whether to use image positional encoding or not makes huge difference on the CROHME 2019 test set. This shows that image positional encoding improves the generalization ability of our model in different scales.

Comparing the \nth{2} and the \nth{3} rows in each dataset, we can see that the model trained using the bidirectional training strategy still outperforms the unidirectionally trained model by about 2.52\% in ExpRate, though both of them using L2R search. This shows that while training a bidirectional language model, the bidirectional training strategy also helps the whole model to extract information from the images more comprehensively. 

Further, using the properties of the bidirectional language model, we evaluate the decoding results of both L2R and R2L directions using approximate joint search, resulting in an improvement of about 4.68\% in ExpRate. 

Finally we report the results using the ensemble method, showing that this can significantly improve the overall recognition performance by about 3.35\% in ExpRate. Specifically, We train five models initialized with different random seeds and average their prediction probabilities at each decoding step.

\begin{table}[htbp]
\renewcommand\arraystretch{1.2}
\caption{Ablation study on the CROHME 2014/2016/2019 test sets(in \%)}
\begin{center}
\begin{tabular}{p{0.15\textwidth}<{\centering}p{0.15\textwidth}<{\centering}p{0.15\textwidth}<{\centering}p{0.15\textwidth}<{\centering}p{0.15\textwidth}<{\centering}p{0.15\textwidth}<{\centering}}
\toprule 
Dataset & IPE & Bi-Trained & AJS  & Ensemble & ExpRate \\
\hline 
\multirow{5}{*}{CROHME14}
& \xmark & \xmark & \xmark & \xmark & $45.13$ \\
& \cmark & \xmark & \xmark & \xmark & $48.17$ \\
& \cmark & \cmark & \xmark & \xmark & $49.49$ \\
& \cmark & \cmark & \cmark & \xmark & $53.96$ \\
& \cmark & \cmark & \cmark & \cmark & $57.91$ \\
\hline
\multirow{5}{*}{CROHME16}
& \xmark & \xmark & \xmark & \xmark & $43.33$ \\
& \cmark & \xmark & \xmark & \xmark & $44.55$ \\
& \cmark & \cmark & \xmark & \xmark & $46.90$ \\
& \cmark & \cmark & \cmark & \xmark & $52.31$ \\
& \cmark & \cmark & \cmark & \cmark & $54.49$ \\
\hline
\multirow{5}{*}{CROHME19}
& \xmark & \xmark & \xmark & \xmark & $20.77$ \\
& \cmark & \xmark & \xmark & \xmark & $44.95$ \\
& \cmark & \cmark & \xmark & \xmark & $48.79$ \\
& \cmark & \cmark & \cmark & \xmark & $52.96$ \\
& \cmark & \cmark & \cmark & \cmark & $56.88$ \\
\bottomrule
\end{tabular}
\label{tb:ablation}
\end{center}
\end{table}

\subsection{Case Study} \label{sec:case_study}
\begin{figure}
	\centering
	\includegraphics[width=1\textwidth]{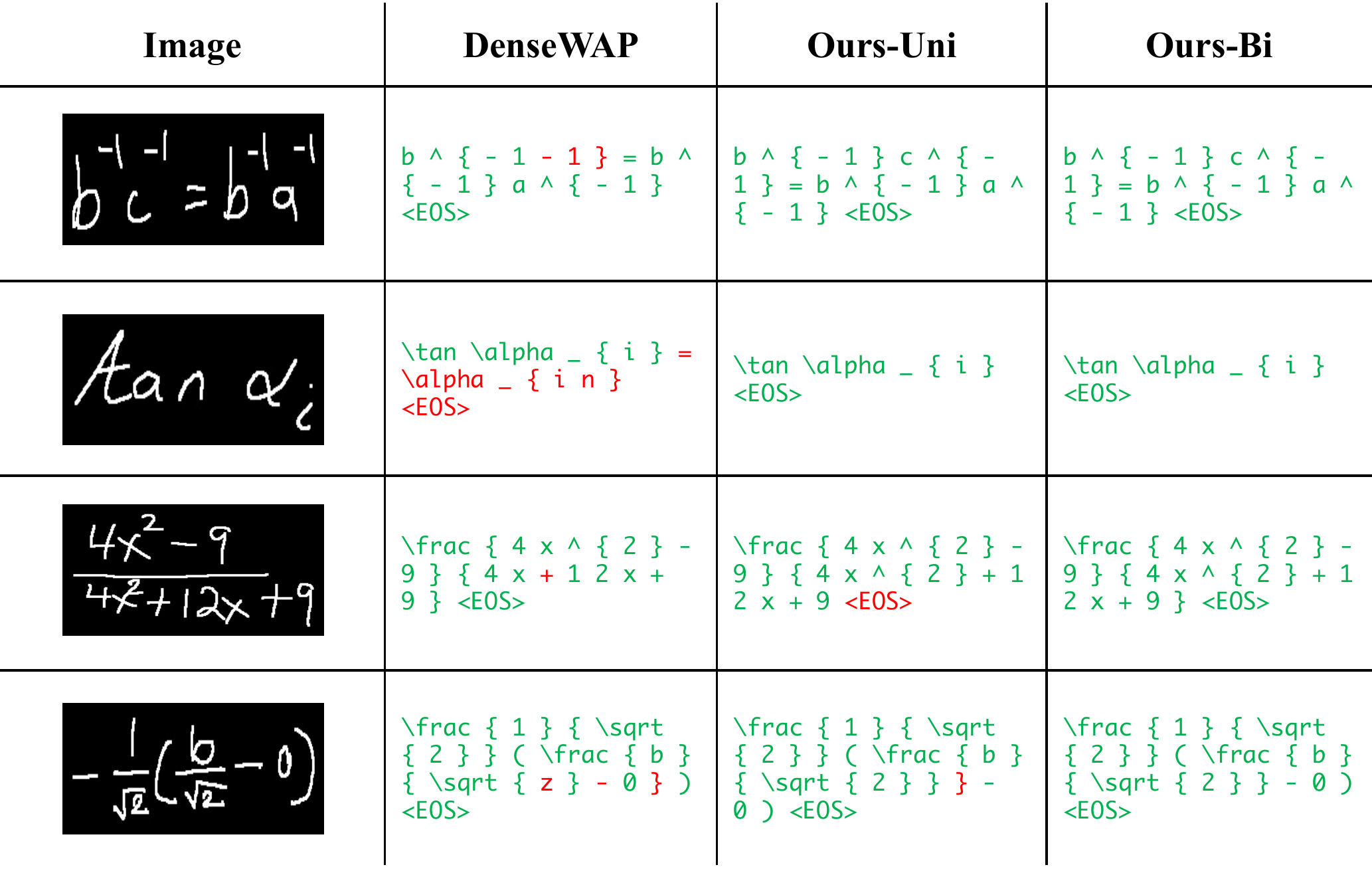}
	\caption{Case studies for the ``DenseWAP"~\cite{zhang2018multi}, ``Ours-Uni" and ``Ours-Bi" models. The red symbols represent incorrect predictions, while the green symbols represent correct predictions.}
	\label{fig:case_study}
\end{figure}

As can be seen in Fig.~\ref{fig:case_study}, we give several case studies for the ``DenseWAP", ``Ours-Uni" and ``Ours-Bi" models. These three models use the same DenseNet encoder. The difference between these three models is that ``DenseWAP" uses an RNN-based decoder, ``Ours-Uni" adapts a vanilla transformer decoder, and ``Ours-Bi" uses the bidirectional training strategy to train transformer decoder.

Firstly, comparing the prediction results between ``DenseWAP" and ``Ours-Uni", we can see that for input images with complex structure, the ``DenseWAP" model cannot predict all the symbols completely. Moreover, for input image with discontinuous structure, the right half is unnecessarily predicted twice. Problems mentioned above reflecting under-parsing and over-parsing phenomenon. However, the ``Ours-Uni" and ``Ours-Bi" models employed with positional encodings are able to identify all the symbols in these images accurately.

Secondly, by comparing the prediction results of ``Ours-Uni" and ``Ours-Bi", we find that ``Ours-Bi" gives more accurate predictions. Owing to the re-score mechanism in approximate joint search procedure, ``Ours-Bi" avoids the asymmetry of ``\{" and ``\}" in the prediction results given by ``Ours-Uni".

\section{Conclusion}
In this paper, a novel bidirectionally trained trans
former model is proposed for handwritten mathematical expression recognition task. Compared to previous approaches, our proposed BTTR model has the following three advantages: (1) Through image positional encoding, the model could capture the location information of the feature-map to guide itself to reasonably assign attention and alleviate the lack of coverage problem. (2) We take the advantage of the transformer model's permutation-invariant property to train a decoder with bidirectional language modeling capability. The bidirectional training strategy enables BTTR model to make predictions in both L2R and R2L directions while ensuring model simplicity. (3) The RNN-based decoder is replaced by the transformer decoder to improve the parallelization of training process. Experiments demonstrate the effectiveness of our proposed BTTR model. Concretely speaking, our model gets ExpRate scores of 57.91\%,  54.49\%, and 56.88\% on the CROHME 2014, CROHME 2016, and CROHME 2019 respectively.

\section*{Acknowledgments}
This work is supported by the projects of National Key R\&D Program of China (2019YFB1406303) and National Natural Science Foundation of China (No. 61876003), which is also a research achievement of Key Laboratory of Science, Technology and Standard in Press Industry (Key Laboratory of Intelligent Press Media Technology).

%
%
%
\bibliographystyle{splncs04}
\bibliography{main}
%
%
%
%
%
\end{document}